\documentclass[letterpaper, 10 pt, journal, twoside]{IEEEtran}

\usepackage{graphics}
\usepackage{amssymb}
\usepackage{pifont}
\usepackage{siunitx}
\usepackage{hyperref}
\usepackage{gensymb}
\usepackage{threeparttable}
\usepackage{arydshln}
\usepackage{enumerate}
\usepackage{float}
\usepackage{lipsum}
\usepackage{dblfloatfix}
\usepackage{float}
\usepackage{caption}
\usepackage{cite}
\usepackage{tikz}

\hypersetup{colorlinks=true}

\newcommand{\circled}[1]{\tikz[baseline=(char.base)]{
            \node[shape=circle,draw,inner sep=0pt, minimum size=0.1cm] (char) {\tiny #1};}}

\title{CEAR: Comprehensive Event Camera Dataset for Rapid Perception of Agile Quadruped Robots}
\author{Shifan Zhu, Zixun Xiong, and Donghyun Kim}

\begin{document}

\twocolumn[{
\renewcommand\twocolumn[1][]{#1}
\maketitle
\vspace{-0.52cm}

\begin{center}
    \centering
    \captionsetup{type=figure}
    \includegraphics[width=\textwidth]{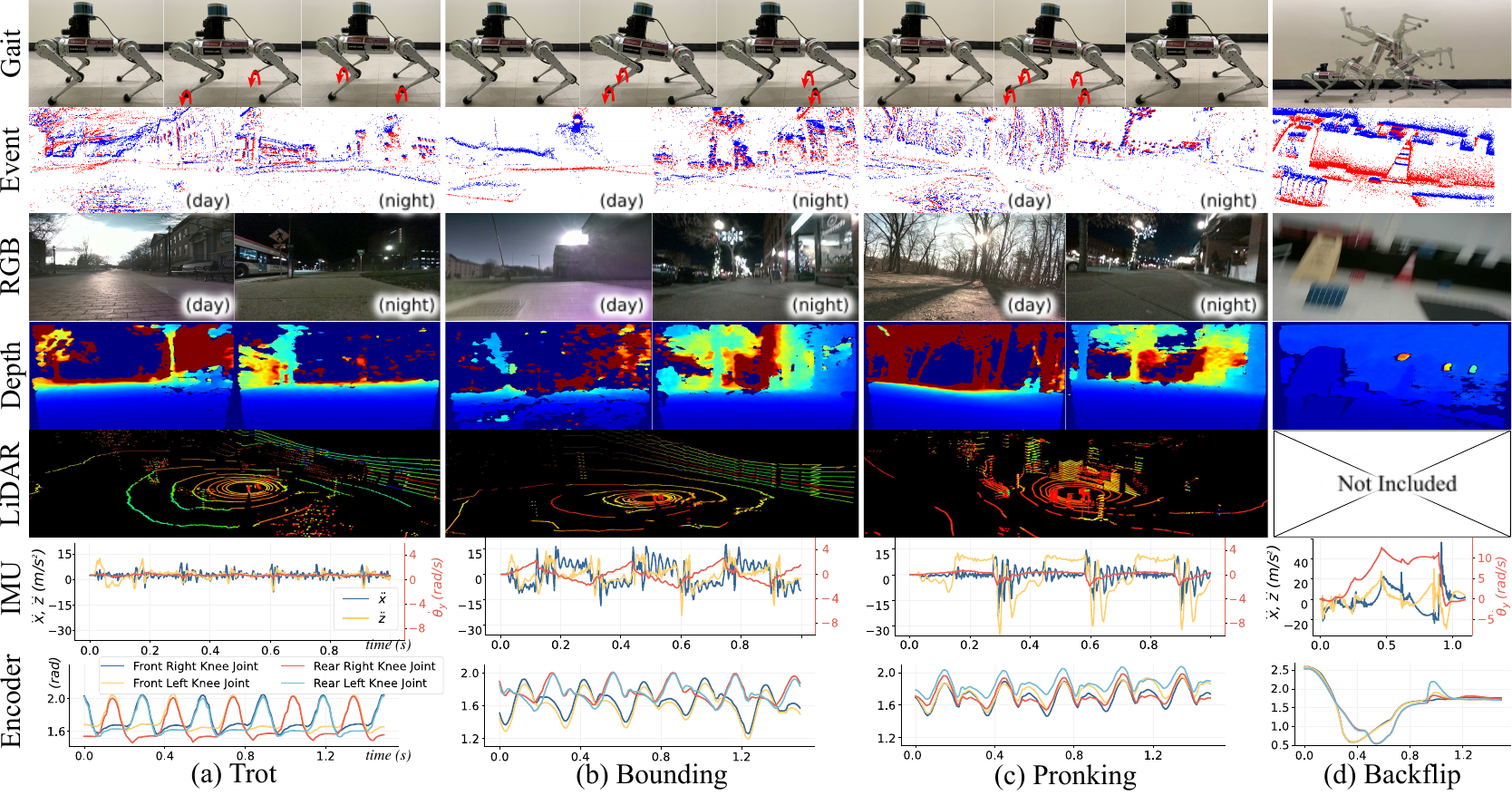}
    \captionof{figure}{\textbf{An overview of the dataset under various gaits.} The top row illustrates three different gaits and a backflip motion. The subsequent rows show different sensor data corresponding to each gait. Ratios of images are adjusted to fit into the figure.}
    \label{fig:abs}
    \vspace{-0.5ex}
\end{center}
}]

{
  \renewcommand{\thefootnote}
    {\fnsymbol{footnote}}
  \footnotetext{Manuscript received: March, 15, 2024; Revised May, 1, 2024; Accepted June 21, 2022. This letter was recommended for publication by Editor Sven Behnke upon evaluation of the Associate Editor and Reviewers’ comments.}
  \footnotetext{All authors are with Manning College of Information \& Computer Sciences,
        University of Massachusetts Amherst, MA, USA. {\tt\small donghyunkim@cs.umass.edu}
        }
  \footnotetext{Digital Object Identifier (DOI): see top of this page.}
}

\begin{abstract}
When legged robots perform agile movements, traditional RGB cameras often produce blurred images, posing a challenge for rapid perception. Event cameras have emerged as a promising solution for capturing rapid perception and coping with challenging lighting conditions thanks to their low latency, high temporal resolution, and high dynamic range. However, integrating event cameras into agile-legged robots is still largely unexplored. To bridge this gap, we introduce CEAR, a dataset comprising data from an event camera, an RGB-D camera, an IMU, a LiDAR, and joint encoders, all mounted on a dynamic quadruped, Mini Cheetah robot. This comprehensive dataset features more than 100 sequences from real-world environments, encompassing various indoor and outdoor environments, different lighting conditions, a range of robot gaits (e.g., trotting, bounding, pronking), as well as acrobatic movements like backflip. To our knowledge, this is the first event camera dataset capturing the dynamic and diverse quadruped robot motions under various setups, developed to advance research in rapid perception for quadruped robots. The CEAR dataset is available at \href{https://daroslab.github.io/cear/}{https://daroslab.github.io/cear/}.
\end{abstract}


\section{INTRODUCTION}

\IEEEPARstart{I}{n} this paper, we introduce a comprehensive dataset developed for rapid perception of agile quadruped robots that can perform dynamic movements in various lighting conditions and environments. The superior mobility of quadruped robots holds significant potential in applications requiring rapid navigation in complex environments. However, despite significant advancement in their locomotion capability~\cite{kim2019highly, margolis2022rapid}, their application remains limited to the scenarios without immediate needs or rapid exploration~\cite{rouvcek2020darpa, jenelten2024dtc}. This constraint delays the broader adoption of the extremely capable legged robotic system to perform hazardous and time-sensitive tasks, such as disaster response, search and rescue, and firefighting~\cite{yoshiike2017development, lindqvist2022multimodality}. One of the most pressing issues is the compromised perception capability during swift movements: motion blur in RGB images and distortion in LiDAR point clouds hinder precise pose estimation and environmental understanding.

\begin{table*}[t]
    \centering
    \caption{Comparison of datasets with event cameras.}
    \begin{tabular}{ccccccc}
         \hline
         Dataset & \begin{tabular}{c} Event \\ Resolution \end{tabular}  & Depth Sensor & IMU & Platform & Environment & Ground-truth Pose \\
         \hline
         \href{http://ci.nst.ei.tum.de/EBSLAM3D/dataset/}{D-eDVS}~\cite{weikersdorfer2014event} & 128px $\times$ 128px & RGB-D & \ding{55} & Handheld & Indoor & MoCap\\
         \href{https://github.com/fbarranco/eventVision-evbench}{evbench}~\cite{barranco2016dataset} & 240px $\times$ 180px & RGB-D & \ding{55} & Wheeled robot & Indoor & Odometer \\
         \href{https://rpg.ifi.uzh.ch/davis_data.html}{Mueggler et al.}~\cite{mueggler2017event} & 240px $\times$ 180px& \ding{55} & \ding{51} & Handheld & Indoor/Outdoor & MoCap\\
         \href{https://daniilidis-group.github.io/mvsec/}{MVSEC}~\cite{zhu2018multivehicle} & 2 $\times$ 346px $\times$ 260px & LiDAR-16 & \ding{51} & Multiple Robots & Indoor/Outdoor & MoCap/Cartographer\\
         \href{https://fpv.ifi.uzh.ch/}{UZH-FPV}~\cite{delmerico2019we} & 346px $\times$ 260px & \ding{55} & \ding{51} & Drone & Indoor/Outdoor & MoCap\\
         \href{https://sites.google.com/view/dgbicra2019-vivid/?pli=1}{ViViD}~\cite{lee2019vivid} & 240px $\times$ 180px & RGB-D/LiDAR-16 & \ding{51} & Handheld & Indoor/Outdoor & MoCap/LeGO-LOAM\\
         \href{https://visibilitydataset.github.io/}{ViViD++}~\cite{lee2022vivid++} & \begin{tabular}{c} 240px $\times$ 180px \\ 640px$\times$480px \end{tabular} & RGB-D/LiDAR-64 & \ding{51} & Handheld/Car & Indoor/Outdoor & \begin{tabular}{c} MoCap/RTK-GPS \\ LeGO-LOAM \end{tabular}\\
         \href{https://dsec.ifi.uzh.ch/}{DSEC}~\cite{gehrig2021dsec} & 2 $\times$ 640px $\times$ 480px & LiDAR-16 & \ding{55} & Car & Outdoor & RTK-GPS\\
         \href{https://ieee-dataport.org/open-access/agri-ebv-autumn}{AGRI-EBV}~\cite{zujevs2021event} & 240px $\times$ 180px & RGB-D/LiDAR-16 & \ding{51} & Wheeled Robot & Outdoor & LeGO-LOAM\\
         \href{https://cvg.cit.tum.de/data/datasets/visual-inertial-event-dataset}{TUM-VIE}~\cite{klenk2021tum} & 2 $\times$ 1280px $\times$ 720px & \ding{55} & \ding{51} & Handheld/Bike & Indoor/Outdoor & MoCap\\
         \href{https://star-datasets.github.io/vector/}{VECtor}~\cite{gao2022vector} & 2 $\times$ 640px $\times$ 480px & RGB-D/LiDAR-128 & \ding{51} & Handheld/Scooter & Indoor & MoCap/ICP\\
         \href{https://m3ed.io/}{M3ED}~\cite{chaney2023m3ed} & 2 $\times$ 1280px $\times$ 720px & LiDAR-64 & \ding{51} & Multiple Robots & Indoor/Outdoor & Faster-LIO\\
         \hline
         \href{https://daroslab.github.io/cear/}{CEAR} (ours) & \begin{tabular}{c} 346px $\times$ 260px \\ 320px $\times$ 240px \end{tabular} & RGB-D/LiDAR-16 & \ding{51} & Agile legged Robot & Indoor, Outdoor & MoCap/Faster-LIO\\
         \hline
    \end{tabular}
    \begin{tablenotes}
          \small
          \item * Extension \textit{-N} represents a N-beam LiDAR and \textit{-D} means depth camera.
        \end{tablenotes}
    \label{tab:comparison}
        \vspace{-2mm}
\end{table*}


Event cameras, inspired by biological vision mechanisms, detect logarithmic intensity changes in images and offer low latency and high temporal resolution~\cite{lichtensteiner2008128x128}, making them exceptionally suited for handling rapid robotic movements~\cite{gallego2020event}. Additionally, their high dynamic range (HDR) effectively addresses challenges in poorly-lit and brightly-lit settings, thereby broadening potential applications for robots~\cite{vidal2018ultimate, rebecq2019high}. Recent studies on event cameras have increasingly gained attention in systems requiring rapid perception, such as drones and autonomous vehicles~\cite{falanga2020dynamic, chen2020event}, yet their use has not been widely recognized within the legged robot community. Once properly configured, these unique sensors can significantly enhance state estimation and terrain perception, greatly expanding the capabilities of legged robots.

While promising, event cameras also have several limitations when used as standalone sensing systems. Due to their operating mechanism of detecting changes in brightness, event data inherently depend on the motion of the camera itself (ego-motion), making it challenging to differentiate between the camera's movement and changes in the external environment. This issue can obscure the clarity of visual data representations, especially when the robot moves slowly or in directions parallel to environmental textures. Overcoming these limitations and enhancing the functionality of agile-legged robots in diverse environments necessitates the development of algorithms for the synergistic integration of multimodal sensors. However, as of 2024, comprehensive datasets are lacking to support this research.

To bridge this gap, we introduce a new dataset collected by a unique sensor suite on an agile quadruped robot, Mini-Cheetah robot~\cite{katz2019mini}. Our system features an event camera, an RGB-D camera, an IMU, a LiDAR, and joint angle sensors. The integration of event and RGB cameras provides the legged robot with enhanced capabilities for both slow and rapid perception. Depth information, crucial for accurate state estimation and navigation, is supplied by the RGB-D camera and LiDAR. The IMU, which can capture high-frequency angular and acceleration data, is inherently immune to disturbances from external dynamic objects. Furthermore, joint angle data can also provide odometry information through forward kinematics.

Our comprehensive dataset includes various distinct robot gaits, including trotting, bounding, and pronking. It also encompasses a wide range of daily indoor and outdoor environments, under varying lighting conditions (e.g., daytime, nighttime, well-lit, dark, under blinking light, and hybrid light setup having bright light sources in dark environments.). This data collection mechanism provides comprehensive scenarios to evaluate the resilience and adaptability of perception systems across various environments.

In summary, the main contributions of this paper are: 1) the inclusion of dynamic motions of a quadruped robot (e.g., trotting, bounding, pronking, and backflipping) to facilitate rapid perception research, 2) comprehensive coverage of the dataset including over 100 sequences gathered in 31 distinct environments under varying lighting conditions and diverse robotic locomotion gaits, and 3) the provision of 6 DoF ground-truth poses from a motion capture system or an advanced SLAM algorithm, along with precise intrinsic, extrinsic, and temporal offset parameters.

\section{RELATED WORK}

In this section, we provide an overview of existing event camera datasets, which are also summarized in Table~\ref{tab:comparison}. The first event camera dataset, introduced in 2014, featured a $128 \times 128$ eDVS event camera~\cite{conradt2009embedded} and an RGB-D camera~\cite{weikersdorfer2014event}. In 2016, \cite{barranco2016dataset} published a dataset using an event camera ($240 \times 180$ pixels) for visual navigation of a mobile wheeled robot. In the following year, \cite{mueggler2017event} released a comprehensive handheld dataset that includes motion speed information as a reference for investigating the capabilities of an event camera ($240 \times 180$ pixels). These datasets have played a crucial role in the introduction of event cameras in research communities.

\begin{figure}
    \centering\includegraphics[width=0.7\linewidth]{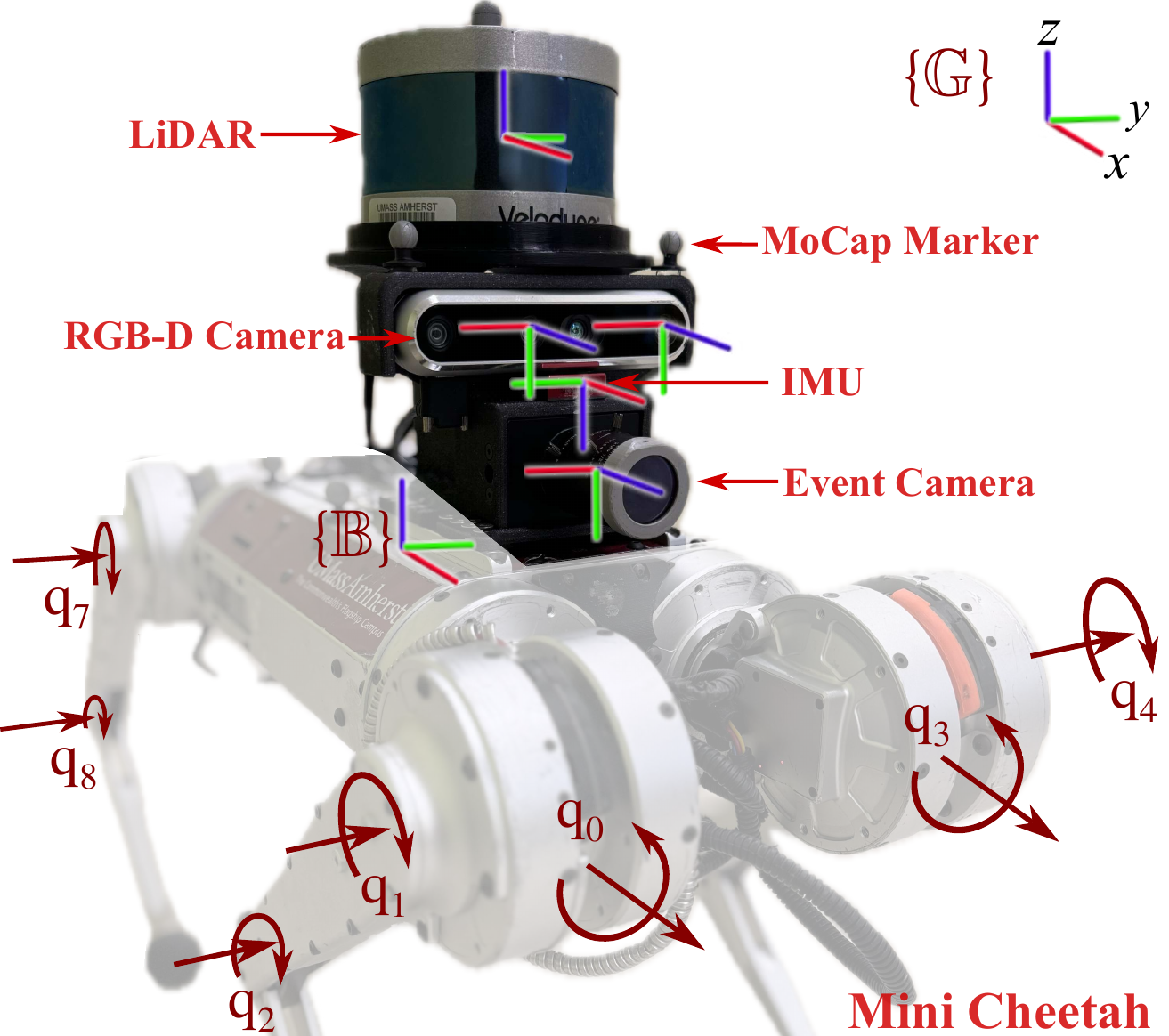}
    \caption{\textbf{Configuration of the sensors.} All sensors are rigidly mounted on top of the Mini Cheetah robot. The coordinate systems for the sensors, the robot's body $\mathbb{B}$, and the global frame $\mathbb{G}$ are indicated by the red, green, and blue axes, representing the $x$, $y$, and $z$ directions, respectively.}
    \label{fig:setup}
    \vspace{-3mm}
\end{figure}

Subsequent developments have led to datasets with higher-resolution event cameras. In 2022, the ViViD dataset~\cite{lee2019vivid} was expanded into ViViD++~\cite{lee2022vivid++}, incorporating a high-resolution event camera ($640 \times 480$ pixels) for driving scenarios. This update included repeated data collection along the predefined trajectories at different times, categorizing motions as slow, unstable, or aggressive. TUM-VIE dataset~\cite{klenk2021tum} includes stereo event camera recordings ($1280 \times 720$ pixels), capturing diverse environments and activities such as walking, running, skating, and biking. Similarly, VECtor dataset~\cite{gao2022vector} features head-mounted and pole-mounted stereo event camera sequences ($640 \times 480$ pixels) in indoor settings. These enhanced-resolution event camera datasets improved image clarity and details, enabling more precise analysis.

In the evolution of event camera datasets, datasets for specific research purposes have also been developed. The DDD17~\cite{binas2017ddd17} and DDD20~\cite{hu2020ddd20} datasets, for instance, provide comprehensive metadata including vehicle speed, GPS positioning, and detailed driving dynamics such as steering, throttle, and brake inputs, improving steering prediction accuracy. \cite{fischer2020event} aimed to tackle the place recognition challenge by mounting a DAVIS346 event camera ($346 \times 260$ pixels) on a car, capturing data along the same trajectory at different times. \cite{zujevs2021event} created agricultural robotics datasets in varied agricultural settings using a mobile wheeled robot. Additionally, \cite{gehrig2021dsec} utilized a dual stereo camera setup with a 16-channel Velodyne LiDAR and an RTK-GPS system for automotive driving scenarios. These datasets delve deeper into specific fields and provide the opportunity to explore the potential of event cameras in focused applications.

Another trend of event camera datasets is their application in various robotic platforms. \cite{delmerico2019we} introduced a dataset focused on aggressive drone racing motions. \cite{zhu2018multivehicle} launched the first multi-robot dataset in 2018, featuring indoor and outdoor places with handheld devices, hexacopters, vehicles, and motorcycles. Additionally, \cite{chaney2023m3ed} unveiled the M3ED dataset, showcasing event cameras across diverse sensor arrays on multiple robotic platforms, including unmanned aerial vehicles, wheeled ground vehicles, and legged robots. 
Although the M3ED dataset explores event cameras with a quadruped robot, Spot, it did not include the dynamic movements of the robot, such as pronking or bounding gaits. The M3ED dataset features relatively conservative trotting gait only and limited environmental variations. On the other hand, our dataset encompasses various, dynamic quadruped robot gait and motions, multi-modal sensor data including dense depth data, and diverse scenarios gathered in distinct environments under different lighting conditions.

\section{Hardware Setup}

Fig.~\ref{fig:setup} illustrates our hardware setup, including an event camera, a RealSense D455 RGB-D camera, a 9-axis VectorNav IMU, a 16-channel Velodyne LiDAR, and 12 joint encoders on the Mini Cheetah robot. We use two different event cameras: a DAVIS346 event camera for outdoor and a DVXplorer Lite event camera for indoor/backflip data collection due to its superior throughput of 100 million events per second (MEPS), compared to the DAVIS346's 12 MEPS. All vision sensors and an IMU are rigidly mounted on the Mini Cheetah robot using custom 3D-printed fixtures. The Mini Cheetah's 12 actuators control its limbs, with joint numbering starting from the front right abduction/adduction joint and continuing through the hip and knee flexion/extension joints, and then extending to the front left, hind right, and hind left limbs as depicted in Fig.~\ref{fig:setup}. Sensor specifications are summarized in Table~\ref{tab:spec}.

\begin{table}
    \caption{Hardware specifications}
    \centering
    \begin{tabular}{ccc}
        \hline
        Sensor Type & Frequency & Specification\\
        \hline
        DAVIS346& N/A & \begin{tabular}{c} Resolution: 346 px $\times$ 260 px \\ FoV: $70\degree \times 56\degree$ \end{tabular} \\
        \hline
        DVXplorer Lite& N/A & \begin{tabular}{c} 320 px $\times$ 240 px \\ FoV: $61\degree \times 52\degree$ \end{tabular} \\
        \hline
        RealSense D455 & 60 & \begin{tabular}{c} Resolution: 640 px $\times$ 480 px \\ RGB FoV: $80\degree \times 65\degree$ \\ Depth FoV: $80\degree \times 64\degree$ \end{tabular}  \\
        \hline
        Velodyne VLP-16 & 10 & \begin{tabular}{c}Range: 100 meters \\ 16 channels \end{tabular} \\
        \hline
        VectorNav VN-100 & 400 & \begin{tabular}{c} 3-axis accelerometer \\ 3-axis gyroscope \\ 3-axis magnetometer \\ 3 DoF orientation \end{tabular}\\
        \hline
        Mini Cheetah Motors & 100 & 12 encoders\\
        \hline
        OptiTrack & 120 & 8 \textit{PrimeX-22} cameras\\
        \hline
    \end{tabular}
    \label{tab:spec}
\end{table}

\subsection{Sensors Overview}
The DAVIS346 event camera ($346 \times 260$ pixels), with a $70\degree$ horizontal and $56\degree$ vertical field of view and a $120$ dB dynamic range, includes a built-in 6-axis IMU and the capability to capture RGB images. The DVXplorer Lite ($320 \times 240$ pixels), featuring a $61\degree$ horizontal and $52\degree$ vertical field of view and a $110$ dB dynamic range, also includes a built-in 6-axis IMU but cannot output RGB images. Both event cameras feature sub-millisecond latency and operate effectively in lighting conditions up to 100k lux, with 50\% of their pixels responding to 80\% contrast. The DVXplorer Lite features a dynamic range of approximately 110 dB and functions from 0.3 lux, while the DAVIS346 offers a slightly higher range of 120 dB starting from 0.1 lux. We put an infrared filter in front of both event cameras to block emissions from the RealSense RGB-D camera.

The RealSense camera D455 operates at $640 \times 480$ resolution and $60~\si{\hertz}$, with $80\degree \times 65\degree$ and $80\degree \times 64\degree$ fields of view for its RGB and depth cameras, respectively. Timestamps are set at the midpoint of exposure, and auto exposure is enabled to obtain well-exposed images in different lighting conditions. Thus, the timestamp gap between RGB and depth frames varies, typically from $0$ to $8.33$~\si{\milli\second}.
VectorNav VN-100 is a 9-axis IMU that comprises a 3-axis gyroscope, 3-axis accelerometer, and 3-axis magnetometer at $400~\si{\hertz}$. Additionally, the dataset also includes the orientation data produced by the IMU sensor in quaternion representation.
The Velodyne VLP-16 LiDAR delivers highly accurate depth measurements as point clouds at $10~\si{\hertz}$. The LiDAR is mounted atop the Mini Cheetah, ensuring an unobstructed 360$\degree$ horizontal and $\pm15\degree$ vertical FoV of the surrounding environment, with a range of up to 100 meters. The Mini Cheetah's 12 custom motors feature joint encoders that measure angular position at $1~\si{\kilo\hertz}$. For this dataset, data is recorded at $100~\si{\hertz}$.

To gather ground-truth pose data, we use the OptiTrack motion capture (MoCap) system, equipped with 8 $\textit{PrimeX-22}$ cameras in a $7.5~\si{\meter} \times 5.5~\si{\meter} \times 2.8~\si{\meter}$ space. This setup is ideal for precise motion tracking and generating accurate 6 DoF ground-truth poses.

\subsection{Calibration}
The calibration process encompasses intrinsic calibration of cameras and an IMU, as well as extrinsic calibration across all sensors. All parameters are available on the dataset website.

\subsubsection{Intrinsic Calibration}
We performed intrinsic parameter calibration of event cameras to empirically measure focal length, optical center, and distortion coefficients. The Kalibr toolbox~\cite{furgale2013unified} is used with the grayscale images constructed from the event stream. We followed standard steps by observing a checkerboard of known dimensions from varying angles. For IMU intrinsic calibration, we used Allan Variance ROS toolbox\footnote{\url{https://github.com/ori-drs/allan_variance_ros}}, involving 6 hours of static data collection for analysis, to identify random walk and noise density of a VectorNav IMU. For the RealSense camera, we did not perform additional intrinsic parameter calibration and used onboard parameters provided by the manufacturer. For confirmation, we qualitatively validated the parameters by checking the alignment between RGB images and the depth images projected into the RGB frame.

\begin{figure}
    \centering\includegraphics[width=\linewidth]{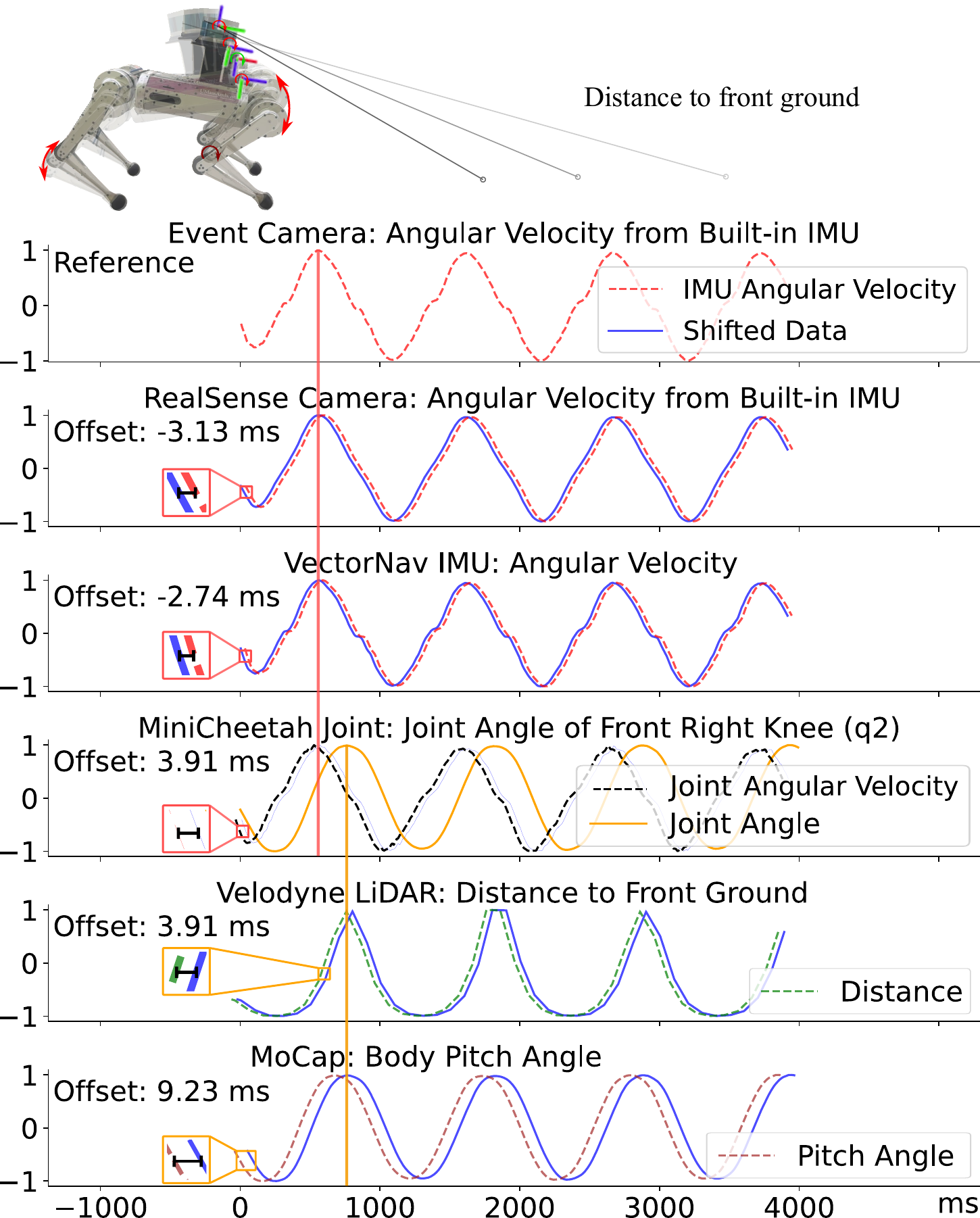}
    \caption{\textbf{Time synchronization across different sensors.} The red vertical line indicates alignment of the RealSense camera and VectorNav IMU, and the event camera's angular velocity, while the orange line shows the alignment of LiDAR, MoCap, and Mini Cheetah's joint angle. For visualization, we magnified the average offset by 10 to distinguish the difference between the original and synchronized data.}
    \label{fig:syn}
    \vspace{-1em}
\end{figure}

\subsubsection{Extrinsic Calibration}

Extrinsic calibration process determines the spatial transformation between each pair of sensors. Due to its high image quality, we used the RealSense RGB camera as a reference to identify extrinsic parameters with other frames -- an event camera, a VectorNav IMU, a LiDAR, and a robot. The Kalibr toolbox was utilized to acquire extrinsic parameters between the RealSense camera, event camera, and the VectorNav IMU, while the Autoware toolkit\footnote{\url{https://github.com/autowarefoundation/autoware}} facilitated the extrinsic calibration between the RealSense RGB camera and LiDAR. Extrinsic parameters between the RealSense RGB camera and the robot were determined from the CAD file. For the RealSense RGB and depth cameras, we used onboard extrinsic parameters provided by the manufacturer. We then derived the extrinsic parameters for all sensor pairs using the identified parameters, creating completed spatial relationships within our sensor array.

\subsection{Time Synchronization}\label{sec:time_synch}
Although every sensor includes accurate internal clock signals, the timestamp of each clock has different offsets. Setting a single timestamp across different sensors is crucial, and hardware synchronization -- a single clock source shared by multiple sensors -- is often used. However, considering the outdoor data gathering, acrobatic backflip, and various sensors (e.g., motion capture and the robot's joint encoders), the hardware connection of all sensors was not a practical option for us. Instead, we introduced a specific motion designed for accurate temporal synchronization through post-processing.

Every time we start the data collection, the Mini-cheetah performs a body pitching swing to produce specific sinusoidal patterns in the sensor data. Since each camera sensor includes a built-in IMU, we compare the angular velocity data from the event and RealSense cameras with the VectorNav IMU, using only the data aligned with the global y-axis. During this process, an event camera is used as a reference and the timestamps of other sensors are offset based on it. Then, the robot's joint encoder data are synchronized with the event camera by comparing the angular velocity of the knee joint and the built-in IMU of event camera. 

\begin{figure}
    \centering\includegraphics[width=\linewidth]{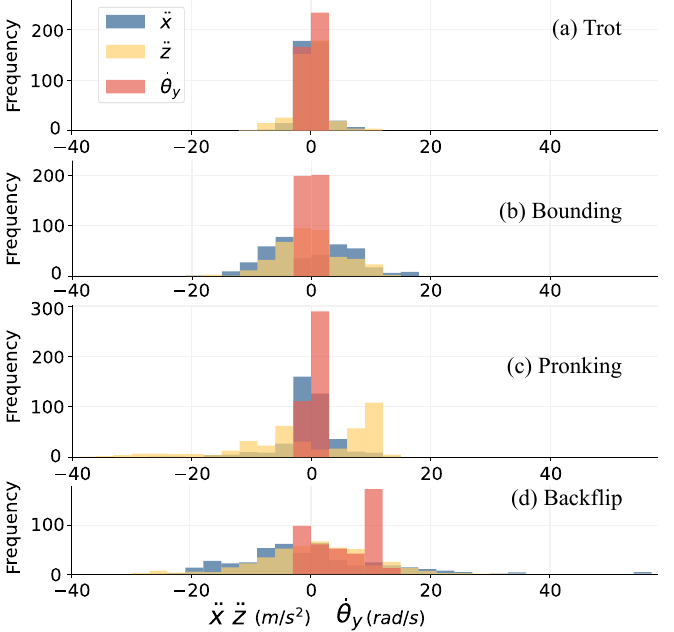}
    \caption{\textbf{Histograms of forward/vertical accelerations, and pitch angular velocity during different gaits.} The histogram depicts the unique features of each gait. Compared to the stable trot gait, the pronking gait exhibits high-dynamic vertical movement, as indicated by high $\ddot{z}$ values. Similarly, the backflip motion shows distinctive high body pitch velocity ($\dot{\theta}_y$) and a wide range of acceleration ($\ddot{x}$, $\ddot{z}$).}
    \label{fig:gaits_hist}
         \vspace{-1em}
\end{figure}
For the LiDAR and MoCap data synchronization, we used joint position data as a reference. Synchronizing the LiDAR sensor is particularly challenging due to the lack of a velocity measurement sensor such as an IMU. To make a sinusoidal pattern in LiDAR data, we perform the pitch swing sequence on flat ground and use the distance from the LiDAR sensor to the front ground as a pattern for matching. By comparing this distance, the robot's knee joint angles, and the body pitch computed from MoCap data, we performed temporal synchronization across all sensors.

In the computation of temporal offset between sensors, we initially normalized all data to standardize measurement scales. We then uniformly sampled 10000 data points from all sensor data by applying linear interpolation to account for the different sampling rates of each sensor. Using these sampled data, we calculated temporal offsets by maximizing cross-correlation. Note that each data sequence includes a body pitch swing motion as well as a ball thrown in front of the robot. This ball data was used to verify the temporal synchronization accuracy across all camera sensors. Fig.~\ref{fig:syn} illustrates the time synchronization results.

\section{Dataset}
The CEAR dataset contains 106 sequences including 50 indoor, 40 outdoor, and 16 backflip sequences. Our dataset aims to evaluate algorithm performance: 1) across diverse quadruped gaits and lighting conditions via indoor and outdoor sequences, and 2) in agile motions through backflip sequences. The name of each sequence reflects the recording environment, lighting conditions, and the gaits of the quadruped robot. For instance, \textit{downtown1\_day\_trot} represents data gathered in a downtown environment during daytime with a trot gait, while \textit{classroom\_blinking\_comb} indicates data collected in a classroom environment under blinking light conditions with a combination of trotting, bounding, and pronking gaits. Fig.~\ref{fig:gaits_hist} further illustrates dynamic characteristics inherent to each gait. In addition, we marked the initial positions of the robot's four feet in each sequence and confirmed its return to these exact positions, providing an additional metric for pose estimation.

\begin{figure*}
    \centering\includegraphics[width=\linewidth]{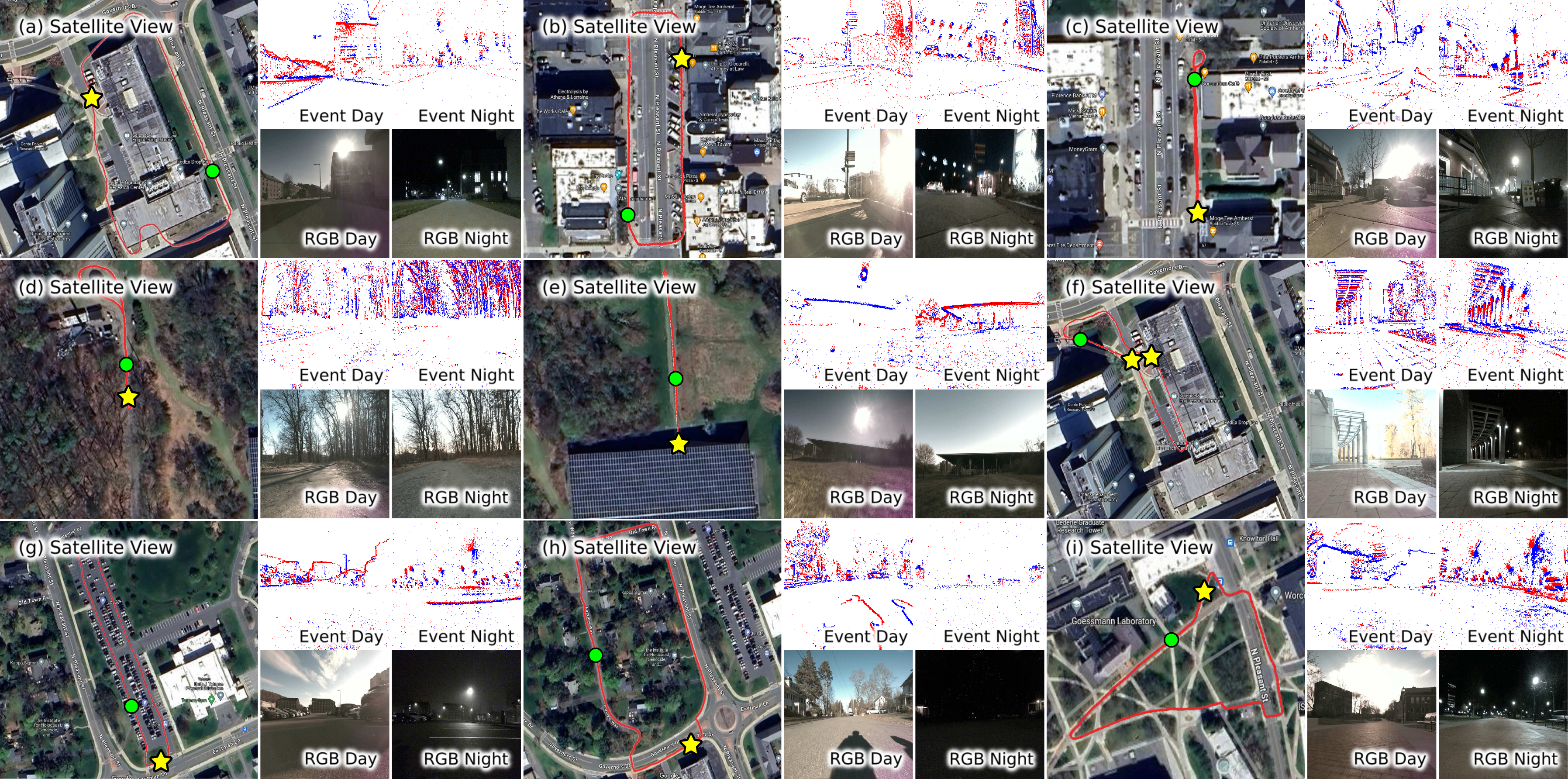}
    \caption{\textbf{Overview of 10 outdoor environments.} (a) is the area around a campus building with a direct sunlight source. (b) and (c) are downtown areas containing dynamic elements like pedestrians, vehicles, and blinking neon lights. (d) is a forest environment that includes slippery ground. (e) is a grassy environment with featureless ground, expansive space, and distant sky. (f) includes two environments, one is a short sidewalk path and the other one is a route between two buildings, highlighting a standard pedestrian space environment. (g) is a parking lot during a cloudy day and a foggy night. (h) is a residential area with complete darkness at night. (i) presents a long sidewalk path, illustrating several pavement types. Event and RGB images are presented with satellite views at each site, showing day and night scenes at the same location, indicated by green circles. The yellow star marks the start/end points, and the red path denotes the trajectory.}
    \label{fig:outdoor_scenario}
    \vspace{-1em}
\end{figure*}

\subsection{Outdoor Sequences}
\begin{figure*}
    \centering\includegraphics[width=\linewidth]{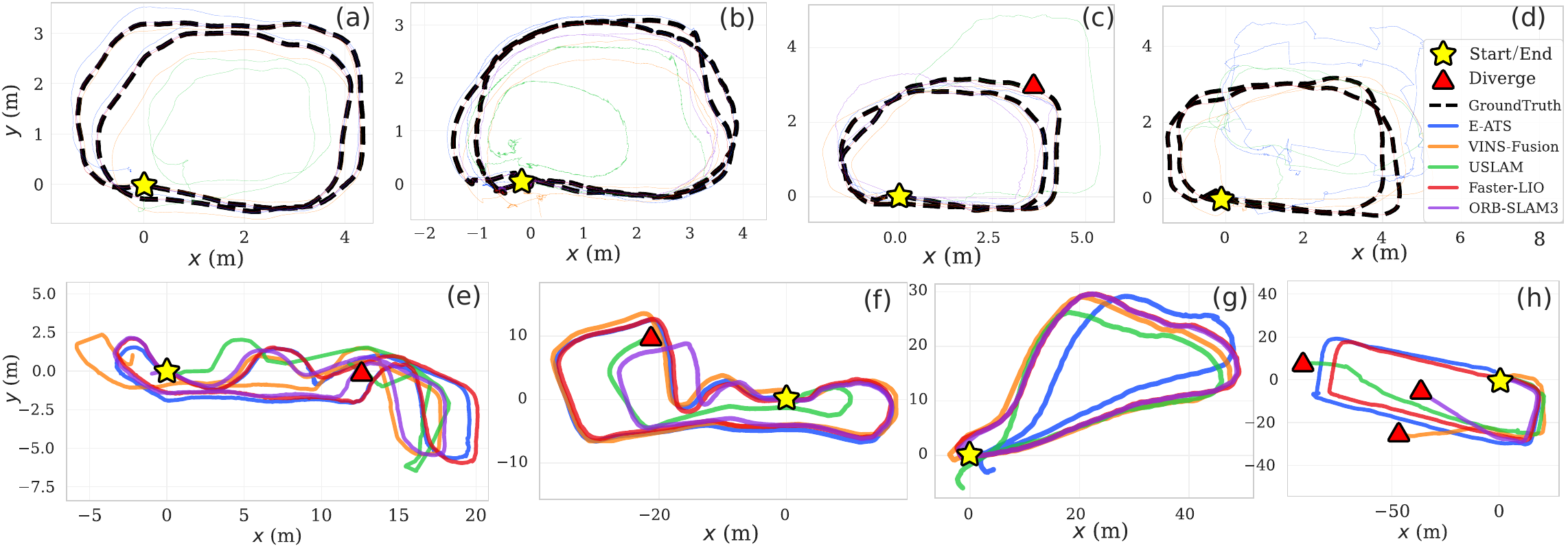}
    \caption{\textbf{Results of various SLAM algorithms on our dataset.} (a), (b), (c), and (d) are within an indoor MoCap environment, featuring a trot gait in a well-lit environment, mixed gaits in a well-lit environment, a trot gait in a dark environment, and a trot gait under blinking lighting condition, respectively. (e) is a lab environment with mixed gaits under an HDR lighting condition. (f) occurs in a public space with a trot gait under a well-lit lighting condition. (g) and (h) are outdoor scenarios, with (g) occurring on a sidewalk with a trot gait during the daytime, and (h) presenting a downtown street with mixed gaits during nighttime. The yellow star marks the starting/ending positions and the red triangle indicates the point of divergence.}
    \label{fig:eva}
\end{figure*}

The outdoor sequences were captured in 10 distinctive outdoor environments, with four sequences per environment. Each set of four sequences includes variations of two different sets of quadruped gaits (trot-only and combined gaits) under different periods of the day (daytime and nighttime), ensuring a comprehensive representation of real-world settings. Daytime sequences feature illumination levels ranging from 40 lux to 80000 lux, while nighttime sequences range from 0.5 lux to 10 lux. Fig.~\ref{fig:outdoor_scenario} provides an illustration of all outdoor environments and associated challenges such as varied light conditions from overexposure to underexposure, sunny, cloudy, and foggy weather, diverse visual patterns including featureless terrains and distant skies, dynamic objects (e.g., vehicles and pedestrians), and slippery surfaces. We include the estimation result of Faster-LIO algorithm~\cite{bai2022faster} as ground truth because of its good matches between the initial and final poses, as shown in Fig.~\ref{fig:eva}, and superior accuracy on indoor MoCap sequences as evidenced in Table~\ref{tab:quantitative}.

\subsection{Indoor Sequences}
The indoor sequences were collected in 13 diverse environments. In the dining hall, building floor, and home environments, we recorded two sequences per site, with each sequence showcasing a different set of quadruped gaits. For other indoor environments where we can control the lighting condition, data collection was expanded to encompass a range of lighting conditions: well-lit (100 lux to 500 lux), dark (0.5 lux to 5 lux), blinking (0.5 lux to 500 lux), and dark environment with bright light sources, which we call high dynamic range (HDR) condition (0.5 lux to 800 lux). 
Specifically, in laboratory environments, four sequences were recorded in one site under both consistently well-lit and HDR lighting conditions. In classroom and kitchen environments, we gathered six sequences in each place across well-lit, HDR, and blinking lighting conditions. In the MoCap space, we arranged various objects on the ground to create three different scenes, recording six sequences in each environment under well-lit, dark, and blinking lighting conditions. Ground-truth poses for the \textit{mocap\_\{$\cdots$\}} sequences are obtained from the MoCap system, while those for other indoor sequences are derived from Faster-LIO. We envision that these indoor sequences will enable a comprehensive evaluation of agile quadruped robots across varied indoor environments and under diverse lighting conditions.

In addition to the challenges similar to outdoor ones such as diverse lighting conditions, dynamic objects, and slippery floors, indoor environments introduce unique complications such as the reflections from floors or glasses that can confuse perception, and closely positioned obstacles that may briefly obstruct cameras' view.

\subsection{Backflip Sequences}
The backflip, involving the rotations up to $750\si{\degree/\second}$, was captured in 7 indoor and 1 outdoor environments, with two sequences in each environment. We excluded the LiDAR sensor due to its substantial weight, which can bother the success of the acrobatic movement, and the significant motion distortion in the point cloud data. Each sequence contains three consecutive backflips and the robot returns to the initial starting position after the three flips. 

Backflip sequences present distinct challenges that are different from other sequences. The backflip motion's high speed leads to significantly blurred images from the RealSense RGB camera. Ground contact at landing causes abrupt changes in acceleration and angular velocity (Fig.~\ref{fig:abs} and Fig.~\ref{fig:gaits_hist}), posing significant challenges for visual-inertial systems and serving as a crucial test for their robustness in agile robot motions. In addition, feature detection becomes also difficult during the acrobatic motion. Although we chose environments that offer visual features from multiple angles, captured images may appear featureless when the camera's view is directed toward the bare ceiling or floor.

\begin{table*}[ht]
\setlength{\tabcolsep}{3pt}
\centering
\caption{Comparison on CEAR Dataset. $\left[\mathbf{R}_{\text{rpe}}:~ ^{\circ}/ \mathrm{s}, \mathbf{R}_{\text{ate}}:~ ^{\circ}, \mathbf{t}_{\text{rpe}}:~ \mathrm{cm} / \mathrm{s}, \mathbf{t}_{\text{ate}}:~ \mathrm{cm}\right]$} \label{tab:quantitative}
\begin{tabular}{lcclcclcclcclcc}
\hline & \multicolumn{2}{c}{ Faster-LIO } & & \multicolumn{2}{c}{ VINS-Fusion } & \multicolumn{3}{c}{ ORB-SLAM3 } & &  \multicolumn{2}{c}{ USLAM } & & \multicolumn{2}{c}{ E-ATS }\\
\cline { 2 - 3 } \cline { 5 - 6 } \cline { 8 - 9 } \cline{ 11 - 12 } \cline{ 14 - 15 }
 & 
$\mathbf{R}_{\text {rpe }} / \mathbf{t}_{\text {rpe }}$ & $\mathbf{R}_{\text {ate }} / \mathbf{t}_{\text {ate }}$ & &
$\mathbf{R}_{\text {rpe }} / \mathbf{t}_{\text {rpe }}$ & $\mathbf{R}_{\text {ate }} / \mathbf{t}_{\text {ate }}$ & &
$\mathbf{R}_{\text {rpe }} / \mathbf{t}_{\text {rpe }}$ & $\mathbf{R}_{\text {ate }} / \mathbf{t}_{\text {ate }}$ & &
$\mathbf{R}_{\text {rpe }} / \mathbf{t}_{\text {rpe }}$ & $\mathbf{R}_{\text {ate }} / \mathbf{t}_{\text {ate }}$ & &
$\mathbf{R}_{\text {rpe }} / \mathbf{t}_{\text {rpe }}$ & $\mathbf{R}_{\text {ate }} / \mathbf{t}_{\text {ate }}$\\
Seq.1 Fig.6a & $1.2 / 3.3$ & $2.4/5.4$ & & $\textbf{0.7} / 4.4$ & $1.4/19.9$ & & $0.8 / \textbf{1.0}$ & $\textbf{0.6/2.5}$ & &
$1.9 / 18.3$ & $16.4/104.6$ & & $0.7 / 1.4$ & $4.3/25.4$\\
Seq.2 Fig.6b & 
$1.5 / 3.2$ & $2.2/\textbf{8.4}$ & &
$0.7 / 3.4$ & $4.4/43.4$ & &
$\textbf{0.6} / 2.2$ & $\textbf{1.3}/17.3$ & &
$1.8^{\circled{1}}$ / $8.3^{\circled{1}}$ & $14.3^{\circled{1}}$/$109.1^{\circled{1}}$ & &
$0.8 / \textbf{1.4}$ & $3.5/24.5$\\
Seq.3 Fig.6c& 
$1.2 / \textbf{3.6}$ & $\textbf{1.3/7.4}$ & &
$\textbf{0.7} / 3.8$ & $1.7/29.5$ & &
$1.4 / 8.1$ & $4.4/31.7$ & &
$3.1^* / 24.0^*$ & $17.5^*/216.0^*$ & &
$2.3 / 3.8$ & $12.3/65.1$\\
Seq.4 & 
$1.4 / 3.8$ & $\textbf{2.1/8.6}$ & &
$\textbf{1.0} / 4.8$ & $4.6/51.5$ & &
$1.3 / 5.4$ & $4.3/28.6$ & &
$2.2^* / 16.1^*$ & $24.9^*/212.0^*$ & &
$1.4 / \textbf{3.6}$ & $9.7/76.1$\\
Seq.5 Fig.6d& 
$1.6 / \textbf{4.1}$ & $\textbf{2.8/6.0}$ & &
$\textbf{1.0} / 9.0$ & $2.4/66.2$ & &
$*$ & $*$ & &
$2.1 / 22.5$ & $18.3/131.5$ &&
$7.9 / 20.4$ & $32.5/176.3$\\
Seq.6 & 
$1.2 / \textbf{3.5}$ & $\textbf{1.4/8.6}$ & &
$\textbf{1.0} / 4.3$ & $3.3/42.4$ & &
$*$ & $*$ & &
$2.6^* / 11.3^*$ & $13.0^*/124.6^*$ & &
$7.6 / 22.4$ & $47.2/191.7$\\
Seq.7 Fig.7 & 
$\times$ & $\times$ & &
$1.0^* / 1.0^*$ & $2.2^*/5.6^*$ & &
$*$ & $*$ & &
$2.8^* / 2.3^*$ & $7.1^*/2.9^*$ & &
$\textbf{0.9}^* / \textbf{0.8}^*$ & $\textbf{0.8}^*/\textbf{1.0}^*$\\
Seq.8 Fig.6e & 
$GT$ & $GT$ & &
$1.4 / 9.3$ & $4.7/212.2$ & &
$\textbf{1.1} / \textbf{5.4}$ & $\textbf{2.4}/\textbf{105.0}$ & &
$1.7^* / 34.2^*$ & $30.5^*/272.3^*$ & &
$2.5 / 6.1$ & $17.4/166.6$\\
Seq.9 Fig.6f & 
$GT$ & $GT$ & &
$1.4 / 9.3$ & $4.7/118.3$ & &
$\textbf{1.1} / \textbf{5.4}$ & $\textbf{2.4}/351.1$ & &
$1.7^* / 34.2^*$ & $30.5^*/612.7^*$ & &
$2.5 / 6.1$ & $17.4/\textbf{90.7}$\\
Seq.10 Fig.6g & 
$GT$ & $GT$ & &
$1.1 / 19.0$ & $2.5/160.3$ & &
$1.1 / \textbf{9.6}$ & $\textbf{1.8}/\textbf{18.3}$ & &
$1.2^{\circled{2}}/ 31.8^{\circled{2}}$ & $9.0^{\circled{2}}/412.8^{\circled{2}}$ & &
$\textbf{1.0} / 17.1$ & $2.4/343.5$\\
Seq.11 Fig.6h& 
$GT$ & $GT$ & &
$0.8^* / 47.8^*$ & $4.5^*/657^*$ & &
$0.8^* / 15.8^*$ & $7.2^*/475^*$ & &
$1.7^{*\circled{2}} / 51^{*\circled{2}}$ & $31^{*\circled{2}}/1184^{*\circled{2}}$ & & 
$\textbf{0.9/9.5}$ & $\textbf{6.1/476.2}$\\
Seq.12 & 
$GT$ & $GT$ & &
$*$ & $*$ & &
$*$ & $*$ & &
$*$ & $*$ & &
$*$ & $*$\\

\hline
\end{tabular}
\begin{tablenotes}
\item From top to bottom, the sequences are as follows: Seq. 1-6 (mocap environment) depicts various gaits under different lighting conditions: trot gait/well-lit, combined gaits/well-lit, trot gait/dark, combined gaits/dark, trot gait/blinking light, and combined gaits/blinking light. Seq. 7 shows a backflip and Seq. 8 is a trot gait in a laboratory with HDR lighting. Seq. 9 is a trot gait in a well-lit learning center. Seq. 10 shows a trot gait on a sidewalk during the daytime. Seq. 11 features combined gaits at night in a downtown setting while Seq. 12 shows combined gaits at night in a residential area. The symbol * indicates a significant divergence, which means that the trajectory diverges by 10\% of the total trajectory length. In this case, we truncated the trajectory and evaluate only the portion of the data before the divergence. The symbol $\times$ means not included. The term $GT$ is used to denote ground truth. RPE for backflip sequences (Seq.7) is assessed at 0.1-second intervals. USLAM operates in three modes: event + constructed frames + IMU (\circled{1}), event + frame + IMU (\circled{2}), or event + IMU (without \circled{1} and \circled{2}).
\end{tablenotes}
\vspace{-1em}
\end{table*}

\section{Evaluation}
To assess the difficulty and quality of the dataset, we tested various sequences with state-of-the-art algorithms for several sensor pairs, including Faster-LIO (LiDAR + IMU)~\cite{bai2022faster}, E-ATS (Event + RGB-D)~\cite{zhu2023event}, UltimateSLAM (Event + RGB + IMU, noted as USLAM)~\cite{vidal2018ultimate}, VINS-Fusion (RGB + IMU)~\cite{qin2019a}, ORB-SLAM3 (RGB-D with loop closure)~\cite{campos2021orb}. All algorithms are tested using open-source implementation provided by the authors, but with CEAR's intrinsic/extrinsic parameters. We diligently tuned the parameters of each algorithm to maximize the performance and found the different sets for indoor and outdoor environments. In the case of USLAM algorithm, we utilize three different setups. When using DVXplorer Lite, which lacks RGB images, we construct grayscale images from event data by utilizing E2VID~\cite{rebecq2019high} (event + constructed frame + IMU mode). When RGB image is available (DAVIS346), we use default setting (event +  frame + IMU mode). When the environment is dark, we use only event and IMU since the quality of constructed images was low (event + IMU mode). In the evaluation, we use the best results among the ones from three modes. 

For sequences captured within the MoCap environment (labeled as \textit{mocap\_\{$\cdots$\}}), we use ground-truth poses from the MoCap system for evaluation. For sequences without a MoCap system, we use the pose estimated by the Faster-LIO algorithm as ground truth. We evaluate each algorithm based on two key metrics: Absolute Trajectory Error (ATE) and Relative Pose Error (RPE). For RPE computation, rotation is measured in degrees per second and translation is measured in centimeters per second, except for backflip sequences where we measure in degrees and centimeters per 0.1 seconds because the duration of one backflip motion is short. ATE is measured in centimeters and degrees to evaluate pose estimation accuracy. We use the initial 50\% of the poses to align trajectories from different coordinate systems and the evaluation is based on the open-source software EVO\footnote{\url{https://github.com/MichaelGrupp/evo}}.
The evaluation results are summarized in Table~\ref{tab:quantitative} to provide a quantitative comparison. 

\begin{figure}
    \centering\includegraphics[width=\linewidth]{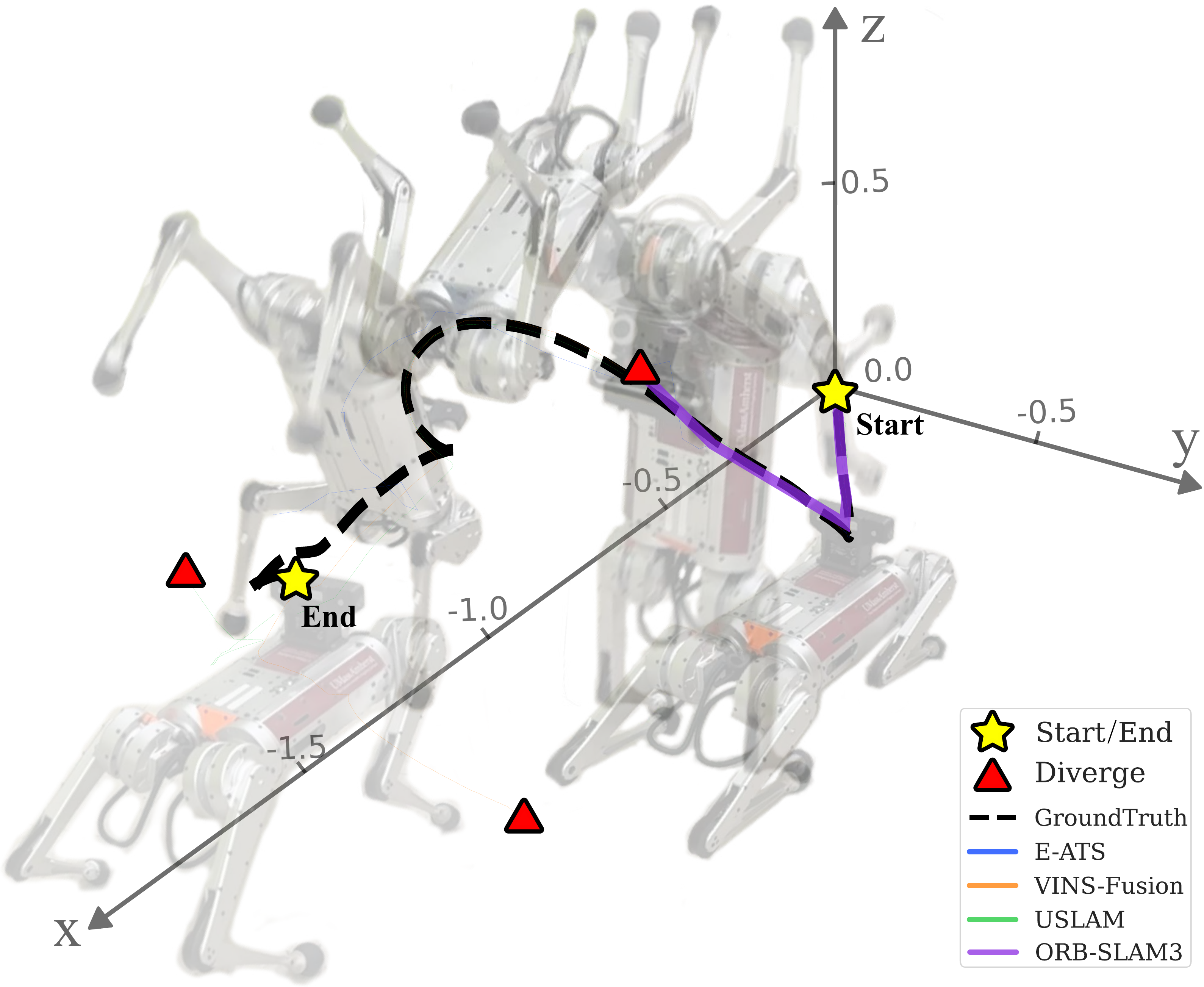}
    \caption{\textbf{Pose estimation during backflip.} This is a subset of env5\_backflip1 that contains three backflips. The yellow star marks start/end positions and red triangle indicates the diverge.}
    \label{fig:backflip3d}
    \vspace{-1em}
\end{figure}

The results show that most algorithms achieve accurate pose estimation with trot gait in well-lit conditions. However, non-event-based algorithms' accuracy drops with dynamic gaits like bounding and pronking, while event-based methods remain stable (Table~\ref{tab:quantitative} Seq.1 and Seq.2). Challenging lighting conditions (dark and blinking) further affect accuracy (Table~\ref{tab:quantitative} Seq.3-6). E-ATS performs poorly in Seq.3-6 due to darkness failing to trigger event data and blinking light breaking the intensity-constant assumption. Notably, backflip sequences, containing three consecutive backflips, present significant challenges, so none of the tested algorithms can handle more than one backflip. Consequently, we limit our evaluation to a single backflip(Fig.~\ref{fig:backflip3d}). Among the four tested algorithms, only one (E-ATS) survived from divergence. Our evaluation results imply that the dataset includes sufficiently challenging scenarios for state-of-the-art algorithms and serve as valuable assets for future benchmarks of advanced event-based visual-inertial odometry algorithms.

\section{CONCLUSIONS}
In this paper, we introduce the CEAR, the pioneering dataset designed to investigate the perception system of a dynamic quadruped robot by utilizing event/RGB-D cameras, LiDAR, IMU, and joint encoders. The CEAR dataset offers a comprehensive collection of indoor and outdoor sequences under various quadruped gaits and lighting conditions. Additionally, it includes backflip sequences to reflect the unique challenges in motion estimation of dynamic legged robots. We envision the CEAR dataset as a cornerstone for exploring the rapid perception of event cameras in robotics research. Furthermore, we anticipate that the dataset will enable the exploitation of unique dynamics introduced by quadruped robots and facilitate the development of specialized algorithms for legged systems. In future work, we plan to add a hardware synchronization module and collect data under more versatile, dynamic movements and develop data-driven methods using CEAR dataset.

\section*{ACKNOWLEDGMENT}
We express our gratitude to Naver Labs and MIT Biomimetic Robotics Lab for providing the Mini-cheetah robot as a research platform.

\bibliographystyle{IEEEtran}
\bibliography{bibliography}

\end{document}